\documentclass[letterpaper]{article} 
\usepackage{aaai25}  
\usepackage{times}  
\usepackage{helvet}  
\usepackage{courier}  
\usepackage[hyphens]{url}  
\usepackage{graphicx} 
\usepackage{natbib}  
\usepackage{caption} 
\usepackage{algorithm}
\usepackage{algorithmic}

\usepackage{newfloat}
\usepackage{listings}
\usepackage{amsmath}
\usepackage[capitalise]{cleveref}
\crefname{table}{Tab.}{Tabs.}
\usepackage{graphicx}
\usepackage{booktabs}
\usepackage{color}
\definecolor{color3}{rgb}{0.95,0.95,0.95}
\usepackage{colortbl}  
\usepackage{xcolor}
\usepackage{amssymb}
\usepackage{multirow}
\usepackage{csquotes}
\DeclareCaptionStyle{ruled}{labelfont=normalfont,labelsep=colon,strut=off} 
\floatstyle{ruled}
\newfloat{listing}{tb}{lst}{}
\floatname{listing}{Listing}
\makeatletter
\renewcommand*{\@fnsymbol}[1]{\ensuremath{\ifcase#1\or * \or \dagger\or \ddagger\or
   \mathsection\or \mathparagraph\or \|\or **\or \dagger\dagger
   \or \ddagger\ddagger \else\@ctrerr\fi}}
\makeatother
\title{MultiBooth:\\
Towards Generating All Your Concepts in an Image from Text}
\author{
    Chenyang Zhu$^{1,}$\thanks{Equal Contribution.},
    Kai Li$^{2,* ,\dagger }$, 
    Yue Ma$^3$,
    Chunming He$^4$,
    Xiu Li$^{1,}$\thanks{Corresponding author.}  \\
}
\affiliations{
    \textsuperscript{\rm 1} Tsinghua Shenzhen International Graduate School, Tsinghua University, Shenzhen, China \\
    \textsuperscript{\rm 2} Meta Platforms, Inc., USA \\
    \textsuperscript{\rm 3}  The Hong Kong University of Science and Technology, Hong Kong \\
    \textsuperscript{\rm 4}  Duke University, Durham, USA \\
    \{chenyangzhu.cs, li.gml.kai, mayuefighting, chunminghe19990224\}@gmail.com,
    li.xiu@mails.tsinghua.edu.cn
}

\usepackage{bibentry}

\begin{document}
\maketitle
\begin{abstract}
This paper introduces MultiBooth, a method that generates images from texts containing various concepts from users.
Despite diffusion models bringing significant advancements for customized text-to-image generation, existing methods often struggle with multi-concept scenarios due to low concept fidelity and high inference cost. MultiBooth addresses these issues by dividing the multi-concept generation process into two phases: a single-concept learning phase and a multi-concept integration phase. During the single-concept learning phase, we employ a multi-modal image encoder and an efficient concept encoding technique to learn a concise and discriminative representation for each concept. In the multi-concept integration phase, we use bounding boxes to define the generation area for each concept within the cross-attention map. This method enables the creation of individual concepts within their specified regions, thereby facilitating the formation of multi-concept images. This strategy not only improves concept fidelity but also reduces additional inference cost. MultiBooth surpasses various baselines in both qualitative and quantitative evaluations, showcasing its superior performance and computational efficiency.
\end{abstract}

%

\section{Introduction}
\label{sec:intro}
The advent of diffusion models~\cite{dalle2,imagen,glide,he2023retidiff,he2024diffusion} has ignited a new wave in the text-to-image (T2I) task, leading to numerous novel methods~\cite{prompt2prompt,ipada,mix,instantid,wang2024cove,wang2024taming}. 
Despite their broad capabilities, users often desire to generate specific concepts such as beloved pets or personal items. These personal concepts are not captured during the training of large-scale T2I models due to their subjective nature, emphasizing the need for customized generation~\cite{ELITE,E4T,facestudio,photomaker,zhu2024instantswap}. Customized generation aims to create new variations of given concepts, including different contexts (e.g., beaches, forests) and styles (e.g., painting), based on just a few user-provided images (typically fewer than 5).

Recent customized generation methods either learn a concise token representation for each subject~\cite{TI} or adopt a fine-tuning strategy to adapt the T2I model specifically for the subject~\cite{DB}. While these methods have achieved impressive results, they primarily focus on single-concept customization and struggle when users want to generate customized images for multiple subjects.
This motivates the study of multi-concept customization (MCC).

\begin{figure*}[tb]
  \centering
  \includegraphics[width=\linewidth]{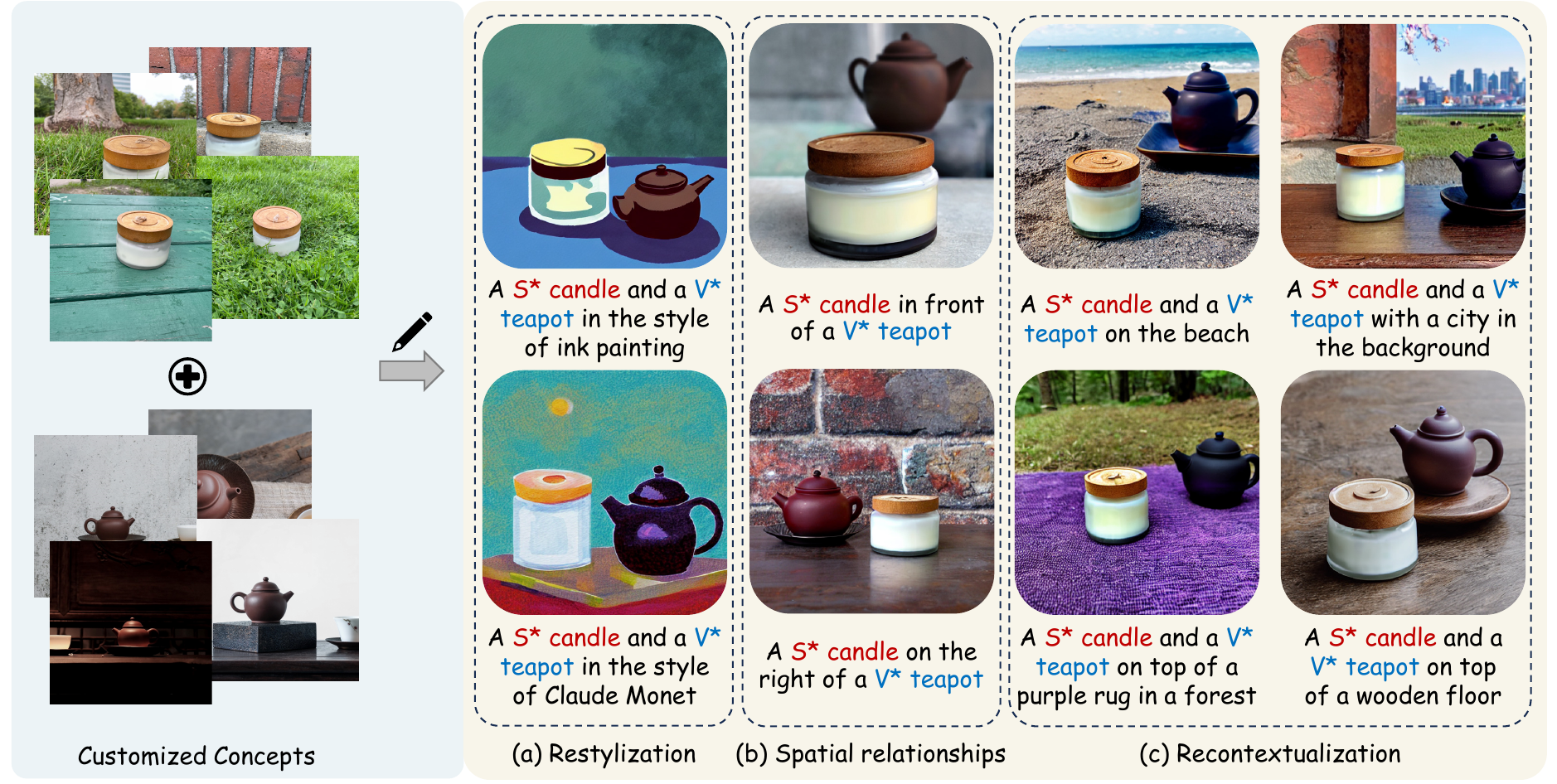}
  \caption{MultiBooth can learn individual customization concepts through a few examples and then combine these learned concepts to create multi-concept images based on text prompts. The results indicate that our MultiBooth can effectively preserve high image fidelity and text alignment when encountering complex multi-concept generation demands, including (a) stylization, (b) different spatial relationships, and (c) contextualization.}
\label{fig:title}
\end{figure*}

Existing methods~\cite{CD} for MCC commonly employ joint training approaches. However, this strategy often leads to feature confusion.
Furthermore, these methods require training distinct models for each combination of subjects and are hard to scale up as the number of subjects grows. An alternative method~\cite{cones} addresses MCC by adjusting attention maps with residual token embeddings during inference. While this approach shows promise, it incurs a notable inference cost. Furthermore, the method encounters difficulties in attaining high fidelity due to the restricted learning capacity of a single residual embedding.

To address the aforementioned issues, we introduce MultiBooth, a two-phase MCC solution that accurately and efficiently generates customized multi-concept images based on user demand.
MultiBooth includes a discriminative single-concept learning phase and a plug-and-play multi-concept integration phase. In the former phase, we learn each concept separately, resulting in a single-concept module for every concept. In the latter phase, we effectively combine these single-concept modules to generate multi-concept images without any extra training.

More concretely, we propose the Adaptive Concept Normalization (ACN) to enhance the representative capability of the generated customized embedding in the single-concept learning phase. We employ a trainable multi-model encoder to generate customized embeddings, followed by the ACN to adjust the L2 norm of these embeddings.
Finally, by incorporating an efficient concept encoding technique, all detailed information of a new concept is extracted and stored in a single-concept module which contains a customized embedding and the efficient concept encoding parameters. 

In the plug-and-play multi-concept integration phase, we further propose a regional customization module to guide the inference process, allowing the correct combination of different single-concept modules for multi-concept image generation. 
Specifically, we divide the attention map into different regions within the cross-attention layers of the U-Net, and each region's attention value is guided by the corresponding single-concept module and prompt. Through the proposed regional customization module, we can generate multi-concept images via any combination of single-concept modules while bringing minimal cost during inference. \cref{fig:title} shows some examples. 


Our approach is extensively validated with various representative subjects, including pets, objects, scenes, etc. 
The results from both qualitative and quantitative comparisons highlight the advantages of our approach in terms of concept fidelity and prompt alignment capability. 
Our contributions are summarized as follows:
\begin{itemize}
    \item We propose a novel framework named MultiBooth. It allows plug-and-play multi-concept generation after separate customization of each concept.
    \item The adaptive concept normalization is proposed in our MultiBooth to mitigate the problem of domain gap in the embedding space, thus learning a representative customized embedding. We also introduce the regional customization module to effectively combine multiple single-concept modules for multi-concept generation.
    \item Our method consistently outperforms current methods in terms of image quality, faithfulness to the intended concepts, and alignment with the text prompts.
\end{itemize}

\begin{figure*}[tb]
  \centering
  \includegraphics[width=\linewidth]{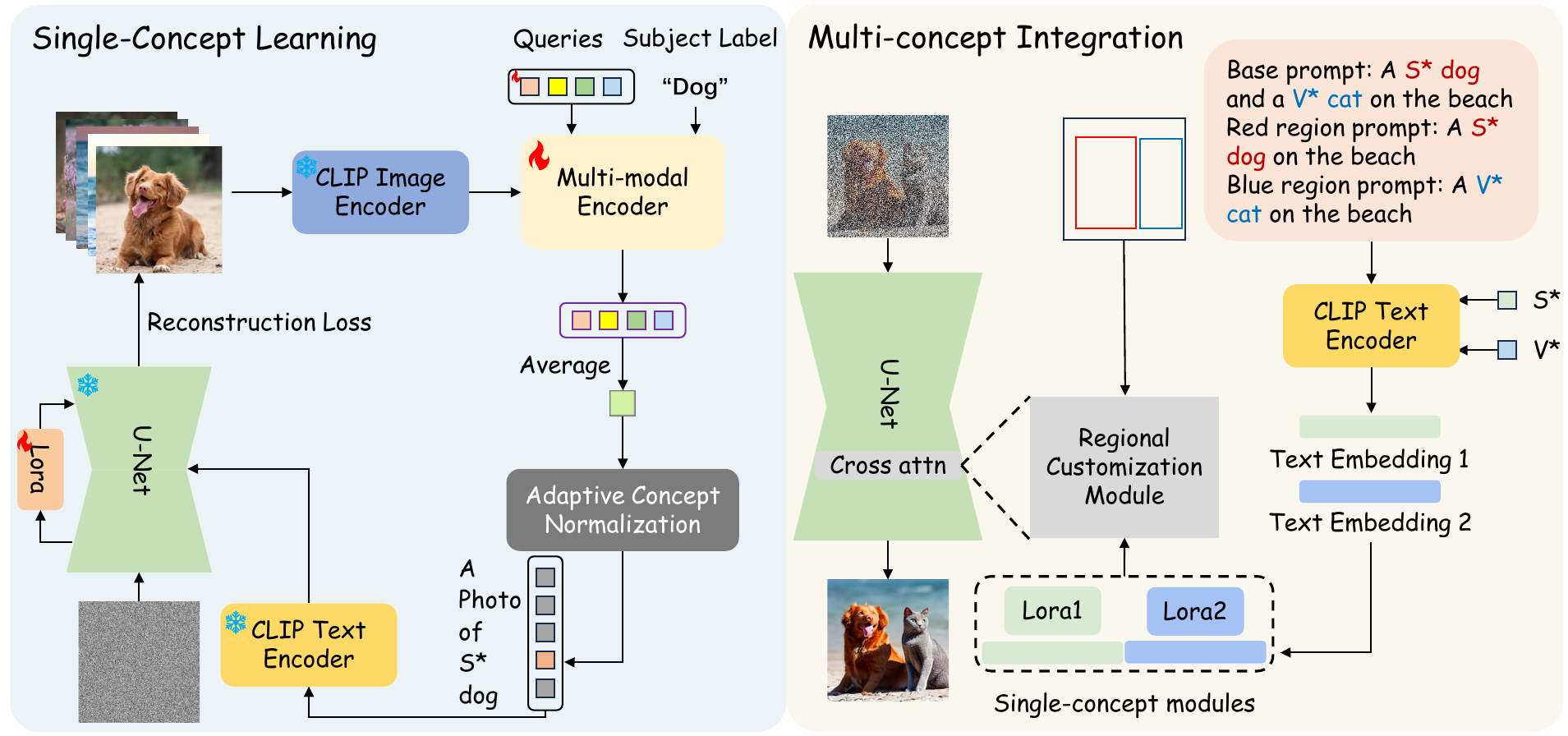}
  \caption{Overall Pipeline of MultiBooth. (a) During the single-concept learning phase, a multi-modal encoder and LoRA parameters are trained to encode every single concept. (b) During the multi-concept integration phase, we first convert $S^*$ and $V^*$ into text embeddings, which are then combined with the corresponding LoRA to form single-concept modules.
  These single-concept modules, along with the bounding boxes, are intended to serve as input for the regional customization module.}
  \label{fig:Framework}
\end{figure*}

\section{Related Work}
\noindent\textbf{Layout-guided text to image generation}. 
T2I models have benefited numerous new tasks~\cite{ma2024followyourpose,ma2024followyourclick,ma2022visual,ma2023magicstick,ma2024followyouremoji,hestrategic,he2023hqg,he2023degradation,fang2024realworld,zhong2024hierarchical,tang2024mind,tang2023consistency,tang2023source,chen2024follow,feng2024dit4edit,wang2024gra,zhong2024goingfeaturesimilarityeffective,zhong2024efficientdatasetdistillationdiffusiondriven}. To achieve finer control that cannot be accomplished using only text prompts, many T2I methods incorporate layout as an additional input to guide the generation process.
One branch of these methods~\cite{boxdiff,attention-refocusing,layout-guidance,directed} involves designing an extra loss function to update the latent variables and guide the sampling process. 
While these methods can achieve image generation in a single forward pass, their fidelity is inadequate when dealing with complex object interactions or attributes.
The other branch of methods~\cite{llm-ground,MultiDiffusion,mixture} performs denoising separately for each layout and subsequently fuses the results, leading to high computational costs.
Different from the aforementioned methods, our method processes all layouts simultaneously, thereby eliminating the need for additional loss functions to guide sampling.
Furthermore, our method can effectively handle complex object interactions while maintaining high image fidelity and precise text alignment.

\noindent\textbf{Customized text to image generation}.
The goal of customized text-to-image generation is to acquire knowledge of a novel concept from a limited set of examples and subsequently generate images of these concepts in diverse scenarios based on text prompts. By leveraging the aforementioned diffusion-based methodologies, it becomes possible to employ the comprehensive text-image prior to customizing the text-to-image process.
The first branch of methods~\cite{TI,disenbooth,cones} achieves customization by creating a new embedding within the tokenizer and associating all the details of the newly introduced concept to this embedding.
The second branch of methods~\cite{ELITE,instantbooth,E4T} trains an adapter to generate embeddings. They need strong GPUs and large datasets for training and only support single-concept customization. To adapt to MCC, they need numerous multi-concept images and costly retraining.
The third branch of methods~\cite{DB,CD} binds the new concept to a rare token followed by a class noun. 
Compared to the previous two branches of methods, they often achieve the best image fidelity. However, this process is achieved by fine-tuning the entire or partial UNet. As a result, they require a larger amount of parameters to store a new concept.
In this work, we utilize a multi-modal model and LoRA to discriminatively and concisely encode every single concept. Then, we introduce the regional customization module to efficiently and accurately produce multi-concept images.

\section{Method}
Given a series of images $\mathcal{S}=\{X_s\}^S_{s=1}$ that represent $S$ concepts of interest, where $\{X_s\} = \{x_i\}^M_{i=1}$ denotes the $M$ images belonging to the concept $s$ which is usually very small (e.g., $M<=5$), the goal of multi-concept customization (MCC) is to generate images that include any number of concepts from $\mathcal{S}$ in various styles, contexts, layout relationship as specified by given text prompts. 

MCC faces significant challenges for two primary reasons. Firstly, learning a concept with a limited number of images is inherently difficult. Secondly, generating multiple concepts \textit{simultaneously and coherently} within the same image while faithfully adhering to the provided text is even harder. To address these challenges, our MultiBooth initially performs high-fidelity learning of a single concept. We employ a multi-modal encoder and the adaptive concept normalization strategy to obtain text-aligned representative customized embeddings. Additionally, the efficient concept encoding technique is employed to further improve the fidelity of single-concept learning. 
To generate multi-concept images, we employ the regional customization module. This module serves as a guide for multiple single-concept modules and utilizes bounding boxes to indicate the positions of each generated concept.
\subsection{Preliminaries}  \label{sec:t2idiff}
In this paper, the foundational model utilized for text-to-image generation is Stable Diffusion~\cite{SD}. 
It takes a text prompt $P$ as input and generates the corresponding image $x$. Stable Diffusion~\cite{SD} consists of three main components: an autoencoder$(\mathcal{E}(\cdot),\mathcal{D}(\cdot))$, a CLIP text encoder $\tau _{\theta}(\cdot)$ and a U-Net $\epsilon_{\theta}(\cdot)$.
Typically, it is trained with the guidance of the following reconstruction loss:
\begin{equation}
\label{loss:rec}
\mathcal{L}_{rec}=\mathbb{E}_{z,\epsilon \sim \mathcal{N}\left( 0,1 \right) ,t,P}\left[ \lVert \epsilon -\epsilon _{\theta}\left( z_t,t,\tau _{\theta}\left( P \right) \right) \rVert_2^2 \right],
\end{equation}
where $\epsilon \sim \mathcal{N}\left( 0,1 \right)$ is a randomly sampled noise, t denotes the time step. The calculation of $z_t$ is given by $z_t=\alpha_t z+\sigma_t \epsilon$, where the coefficients $\alpha_t$ and $\sigma_t$ are provided by the noise scheduler. 

Given $M$ images $\{X_s \} = \{ x_i \}_{i=1}^M$ of a certain concept $s$,
previous works~\cite{TI,DB,CD} associate a unique placeholder string $S^*$ with concept $s$ through a specific prompt $P_s$ like \enquote{a photo of a $S^*$ dog}, with the following finetuning objective:
\begin{equation}
\mathcal{L}_{bind}=\mathbb{E}_{z=\mathcal{E}(x), x\sim X_s,\epsilon ,t,P_s}\left[ \lVert \epsilon -\epsilon _{\theta}\left( z_t,t,\tau _{\theta}\left( P_s \right) \right) \rVert_2^2 \right].
\label{loss:bind}
\end{equation}
Minimizing~\cref{loss:bind} can encourage the U-Net $\epsilon_{\theta}(\cdot)$ to accurately reconstruct the images of the concept $s$, effectively binding the placeholder string $S^*$ to the concept $s$.

\subsection{Single-Concept Learning} \label{sec:single}
\subsubsection{Multi-modal Concept Extraction.}
Existing customization methods~\cite{E4T,ELITE} mainly utilize a single image encoder to encode the whole image into concept embeddings. However, the single image encoder may also encode unrelated objects in the images. 
To remedy this, we employ a multi-modal encoder that takes as input both the images and the concept name (e.g., \enquote{dog}) to generate concise and discriminative customized embeddings.

Inspired by MiniGPT4~\cite{minigpt} and BLIP-Diffusion~\cite{blip-diff}, we utilize the QFormer, a light-weighted multi-modal encoder, to generate the customized embeddings for each concept.
As shown in the left part of~\cref{fig:Framework}, the QFormer encoder $E$ has three types of inputs: visual embeddings $\xi$ of an image, text description $l$ of the concept of interest, and learnable query tokens $W=[w_1,\cdots ,w_K]$ where $K$ is the number of query tokens. 
Given an image $x_i\in X_s$ of concept $s$, we employ a frozen CLIP~\cite{clip} image encoder to extract the visual embeddings $\xi$ of the image.
Subsequently, we set the input text $l$ as the concept name for the image. The learnable query tokens $W$ interact with the text description $l$ through a self-attention layer and with the visual embedding $\xi$ through a cross-attention layer. This interaction results in text-image aligned output tokens $O=E(\xi , l , W)$ with the same dimensions as $W$. Finally, we average these tokens and get initial customized embedding $v_i=\frac{1}{K}\cdot \sum_{i=1}^K{o_i}$.

After obtaining the customized embedding $v_i$ of concept $s$, we introduce a placeholder string $S^*$ to represent the concept $s$, with $v_i$ representing the word embedding of $S^*$.
Through this placeholder string $S^*$, we can easily activate the customized word embedding $v_i$ to reconstruct the input concept image $x_i$ with prompts like \enquote{a photo of a $S^*$ dog}.

\begin{table}[t]
\centering
\small

 \tabcolsep=4pt
 \fontsize{10pt}{9pt}\selectfont
\resizebox{\linewidth}{!}{
\begin{tabular}{l|c c c c c c c c c c c}
\toprule
    \textbf{Method} & \textbf{a} & \textbf{S*} & \textbf{dog} & \textbf{and} & \textbf{a} & \textbf{V*} & \textbf{cat} & \textbf{on} & \textbf{the} &\textbf{beach} \\
    \midrule \midrule
    Textual Inversion & 0.35 & \textbf{2.85} & - & 0.34 & 0.35 & \textbf{0.94} & - & 0.34 & 0.34 & 0.37  \\
    Ours w/o ACN& 0.35 & \textbf{2.35} & 0.37 & 0.34 & 0.35 & \textbf{3.14} & 0.37 & 0.34 & 0.34 & 0.37 \\
    Ours w/o ACN\&Reg& 0.35 & \textbf{111.02} & 0.37 & 0.34 & 0.35 & \textbf{131.24} & 0.37 & 0.34 & 0.34 & 0.37 \\
    \midrule
    Ours  & 0.35 & \textbf{0.37} & 0.37 & 0.34 & 0.35 & \textbf{0.37} & 0.37 & 0.34 & 0.34 & 0.37\\
\bottomrule
 \end{tabular}}
 \caption{Quantization results of the L2 norm of each word embedding in the prompt. 
}
 \vspace{-0.5cm}
 \label{tab:autonorm}
 \end{table}

\begin{figure*}[tb]
  \centering
  \includegraphics[width=\linewidth]{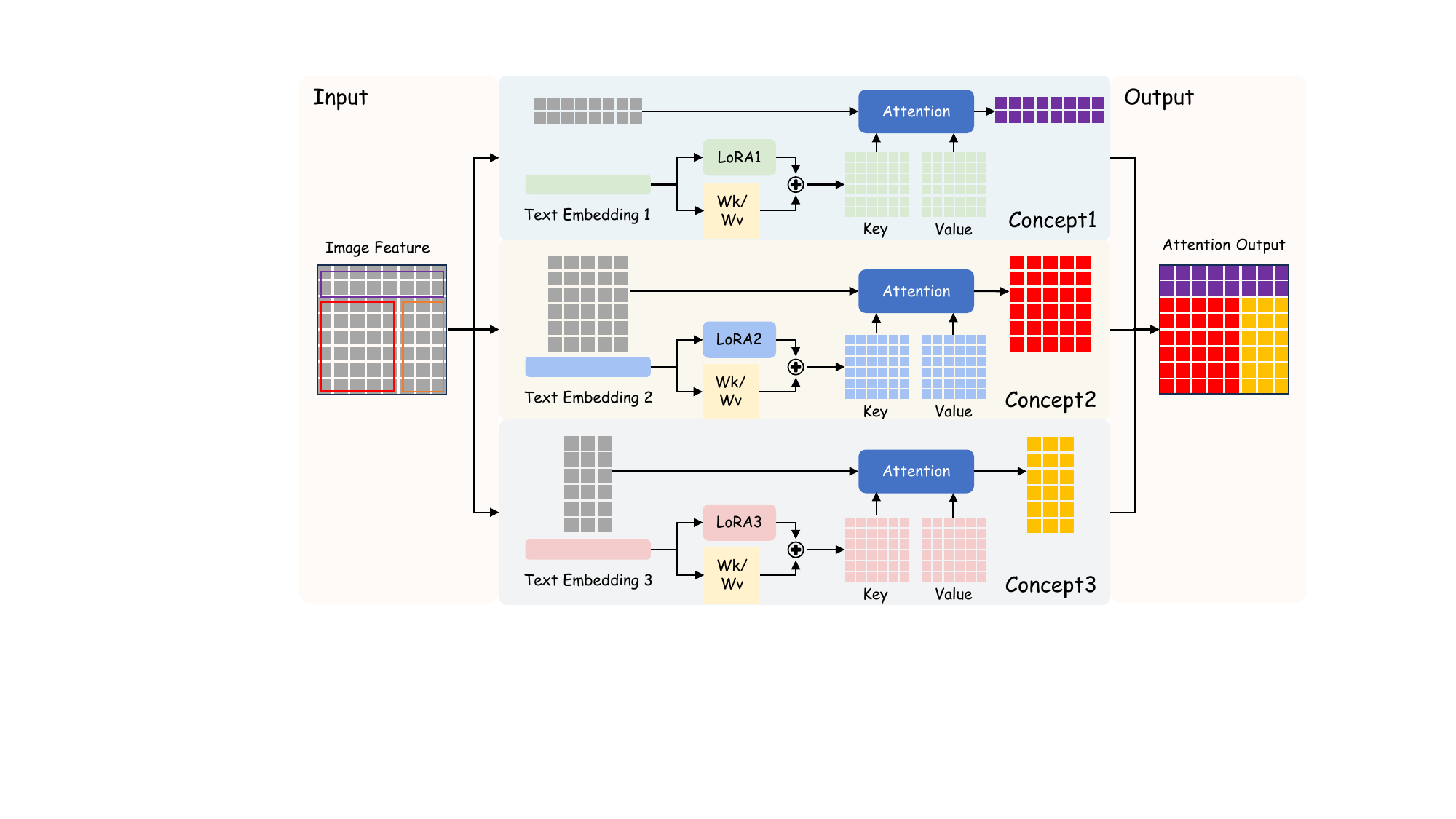}
  \caption{Regional Customization Module. We initially divide the image feature into several regions via bounding boxes to acquire the query $Q$ for each concept.
  Subsequently, we combine the single-concept module with $W_k$ and $W_v$ to derive the corresponding key $K$ and value $V$. After that, we perform the attention operation on the obtained $Q$, $K$, and $V$ to get a partial attention output. The above procedure is applied to each concept simultaneously, forming the final attention output.}
  \label{fig:RCM}
\end{figure*}

\subsubsection{Adaptive Concept Normalization.}
We have observed a domain gap between our customized embedding $v_i$ and other word embeddings in the prompt. As shown in~\cref{tab:autonorm}, the L2 norm of our customized embedding is considerably larger than that of other word embeddings in the prompt.
Notably, these word embeddings, belonging to the same order of magnitude, are predefined within the embedding space of the CLIP text encoder $\tau_\theta(\cdot)$.
This significant difference in quantity weakens the model's ability of multi-concept generation.
To remedy this, we further apply the Adaptive Concept Normalization (ACN) strategy to the customized embedding $v_i$, adjusting its L2 norm to obtain the final customized embedding $\hat{v_i}$.

Our ACN consists of two steps. The first step is L2 normalization, adjusting the L2 norm of the customized embedding $v_i$ to $1$. The second step is adaptive scaling, which brings the L2 norm of $v_i$ to a comparable magnitude as other word embeddings in the prompt. Specifically, let $c_{l} \in \mathbb{R}^d$ represent the word embedding corresponding to the subject name of $v_i$ (e.g., the word embedding of \enquote{dog}), where $d$ is the dimension of embeddings. The adaptive concept normalization $\hat{v_i} =v_i\cdot \frac{\lVert c_l \rVert _2}{\lVert v_i \rVert _2}$.
As shown in~\cref{tab:autonorm}, this operation effectively addresses the problem of domain gap in the embedding space.

\subsubsection{Efficient Concept Encoding.}
To further improve the concept fidelity during single-concept learning and avoid language drift caused by finetuning the U-Net, we incorporate the LoRA technique~\cite{LoRA,he2024weakly} for efficient concept encoding.
Specifically, we incorporate a low-rank decomposition to the key and value weight matrices of attention layers within the U-Net $\epsilon_{\theta}(\cdot)$. Each pre-trained weight matrix $W_{init}\in \mathbb{R}^{d\times k}$ of the U-Net $\epsilon_{\theta}(\cdot)$ is utilized in the forward computation as follows:
\begin{equation}
    h=W_{init}x+\Delta Wx=W_{init}x+BAx,
\end{equation}
where $A \in \mathbb{R}^{r\times k},\ B \in \mathbb{R}^{d \times r}$ are trainable parameters of efficient concept encoding, and the rank $r \ll \min (d,k)$.
During training, the pre-trained weight matrix $W_{init}$ stays constant without receiving gradient updates.
We also use a regularization term to lower the L2 norm of $v_i$ before ACN.
Without this term, the L2 norm of $v_i$ can grow large as shown in \cref{tab:autonorm}. Scaling $v_i$ with ACN could greatly alter its magnitude, causing information loss.
As a result, the whole single-concept learning framework can be trained as follows:
\begin{equation}
\mathcal{L}=\mathbb{E}_{z=\mathcal{E}(x), x\sim X_s,\epsilon ,t,P_s}\left[ \lVert \epsilon -\epsilon _{\theta}\left( z_t,t,\tau _{\theta}\left( P_s \right) \right) \rVert_2^2 \right]+\lambda \lVert {v_i} \rVert_2^2,
    \label{loss:final}
\end{equation}
where $\lambda$ denotes a balancing hyperparameter and is consistently set to 0.01 across all experiments. 

So far, we can learn a new concept efficiently and store its information in a dedicated single-concept module.
This module contains a customized embedding along with the corresponding LoRA parameters. 
The extra parameter for a new concept is less than 7MB, which is significantly lower compared to 3.3GB in DreamBooth~\cite{DB} and 72MB in Custom Diffusion~\cite{CD}.
Furthermore, the single-concept module is plug-and-play for multi-concept generation, as users can combine any single-concept module through the Regional Customization Module to perform multi-concept generation.

\begin{figure*}[tb]
  \centering
  \includegraphics[width=\linewidth]{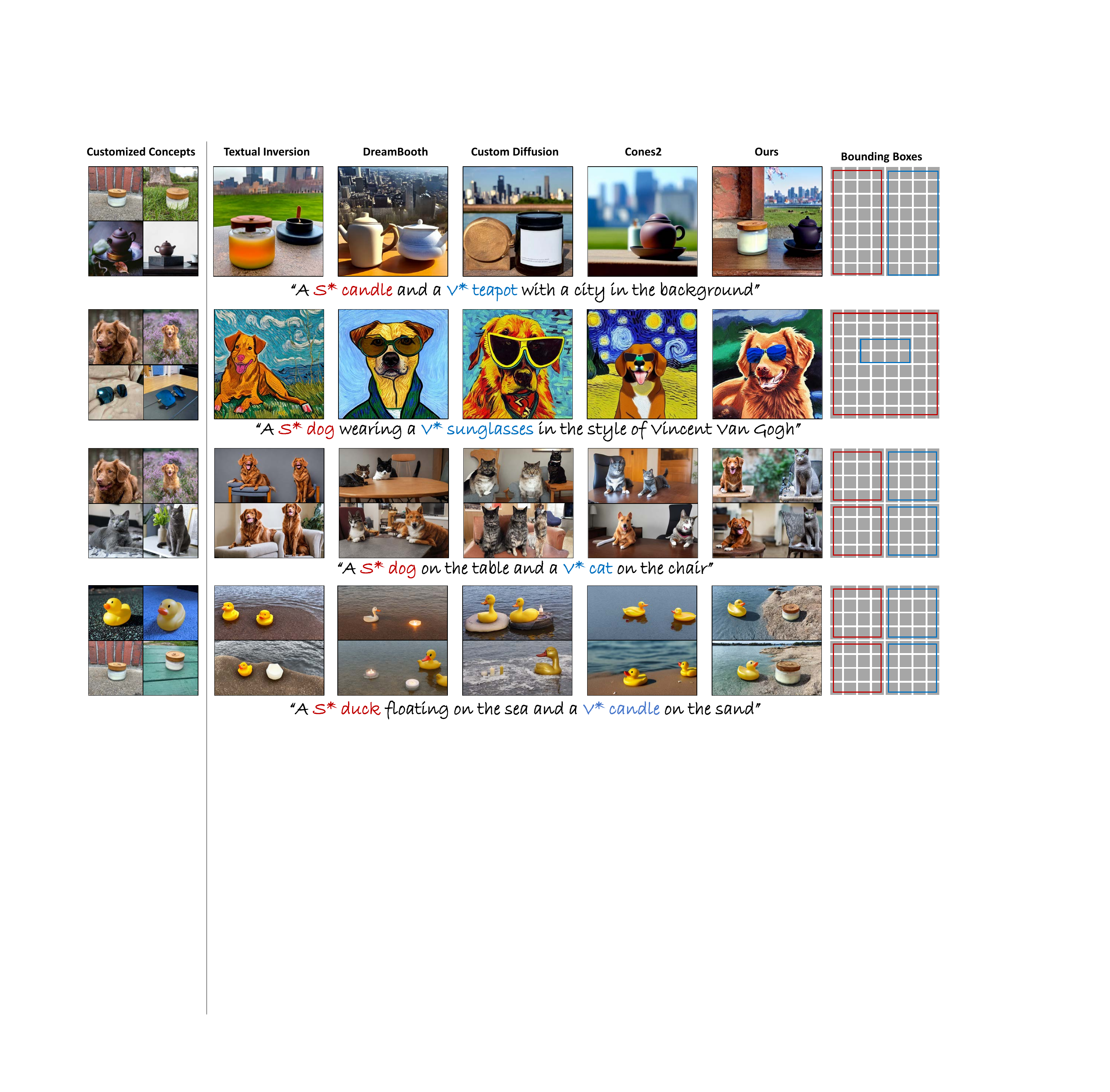}
  \vspace{-0.2cm}
  \caption{Qualitative comparisons. Our method outperforms all the compared methods in image fidelity and prompt alignment. 
  }
  \vspace{-0.5cm}
  \label{fig:main comparison results}
\end{figure*}

\subsection{Multi-Concept Integration} \label{sec:multi}

\subsubsection{Regional Customization Module.}

To integrate multiple single-concept modules for multi-concept generation, we propose the Regional Customization Module (RCM) in cross-attention layers.
The key insight of our RCM is to generate each concept within the specified region and allow different concepts to interact accurately in overlapping regions.

As shown in the right part of~\cref{fig:Framework}, given a base prompt $p_{base}$ describing the desired generated results, we can obtain the bounding boxes $B=\{ b_i \}_{i=1}^{S}$ and the corresponding region prompts $P_r=\{ p_i \}_{i=1}^{S}$ for each concept either through user-defined methods or automated processes (see~\cref{sec:discuss}).
The region prompt guides the concept generation within each specific region, while the base prompt ensures interaction among concepts across different regions.
As a result, the text embeddings $C=\{ c_i \}_{i=1}^{S}$ for each region can be acquired through the combination of the region prompt and the base prompt:
\begin{equation}
c_i=\tau_\theta(p_i)+\tau_\theta(p_{base}),i=1,2,\cdots, S,
\label{equ:base prompt}
\end{equation}
where $c_i\in \mathbb{R}^{k\times d}$, $k$ is the the maximum length of input words and $\tau_\theta(\cdot)$ is the CLIP text encoder.

Then, we integrate the text guidance from text embeddings and the concept information in LoRA into each region \textit{simultaneously} within the cross-attention layers.
As shown in~\cref{fig:RCM}, the image feature $F \in \mathbb{R}^{h\times w}$ is the input of RCM.
For the $i^{th}$ concept, the image feature $F$ is cropped using the bounding box $b_i \in \mathbb{R}^{h_i \times w_i}$, resulting in the partial image feature $f_i \in \mathbb{R}^{h_i\times w_i}$. 
With $f_i$, we can obtain the query vector $Q_i$ through $Q_i=W_q\cdot f_i$.
Next, we derive the key and value vector $K_i$ and $V_i$ using the text embedding $c_i$ and corresponding LoRA parameters $\{A_{ij}, B_{ij}\}_{i=1}^{S}$ through:
\begin{gather}
	K_i=W_k\cdot c_i+B_{i1}A_{i1}\cdot c_i,\\
	V_i=W_v\cdot c_i+B_{i2}A_{i2}\cdot c_i,
\end{gather}
where $A_{ij} \in \mathbb{R}^{r\times k}$ and $B_{ij} \in \mathbb{R}^{d \times r}$, $j=1$ and $j=2$ indicating the low-rank decomposition of $W_k$ and $W_v$ respectively.
In order to derive the text-aligned image feature with concept information, we then apply the attention operation to the query, key, and value vectors:
\begin{equation}
\operatorname{Attn}\left(Q_i, K_i, V_i\right)=\operatorname{Softmax}\left(\frac{Q_i K_i^T}{\sqrt{d^{\prime}}}\right) V_i,
\end{equation}
where $d^\prime$ represents the output dimension of key and query features. 
The image feature $\hat{f_i}=\operatorname{Attn}\left(Q_i, K_i, V_i\right) \in \mathbb{R}^{h_i\times w_i}$ contains both the text guidance and concept information through the attention mechanism and retains its original dimensions. 
For overlapping regions, we use a weighted average strategy to ensure the generation of each concept:
\begin{equation}
\begin{aligned}
    \hat{f}=\frac{1}{\eta}\cdot \sum_{i=1}^\eta{w_i\cdot \hat{f}_i},\ \sum_{i=1}^\eta{w_i}=1,\ \bigcup_{i=1}^{\eta}{b_i}\ne \varnothing, 
\end{aligned}
\end{equation}
where $\eta$ is the number of overlapping concepts, $\hat{f}$ is the output feature of the overlapping region, $w_i$ is the average weight of the $i^{th}$ concept. The setting of $w_i$ is further discussed in $Suppl$.

Compared to~\cite{CD,cones}, our RCM offers more flexible and precise customization that cannot be achieved solely through text prompts.
Once the single-concept modules are obtained, RCM can combine multiple single-concept modules in a plug-and-play manner to perform multi-concept generation without retraining.
With bounding boxes indicating the regions of the generated concepts, RCM can generate each concept according to different region prompts (see~\cref{sec:Qualitative}) and handle complex object interactions under the guidance of the base prompt (see~\cref{sec:discuss}).
Moreover, despite the superior multi-concept customization performance achieved by our RCM, it incurs minimal cost during inference. This is because the RCM generates all the customized concepts \textit{simultaneously}, rather than \textit{sequentially}, which is further discussed in~\cref{sec:discuss}.
We also provide a thorough comparison between our RCM and other layout T2I methods~\cite{llm-ground,boxdiff}, detailed in~\cref{sec:ablation}.

\section{Experiment}
\noindent\textbf{Implementation details.}
All of our experiments are based on Stable Diffusion v1.5 and are conducted on one RTX3090. We set the rank of LoRA to be 16.
During training, we randomly select text prompts $P_s$ from the CLIP ImageNet templates~\cite{clip} following the Textual Inversion~\cite{TI}. 
During training, we optimize for 900 steps with a learning rate of $8 \times 10^{-5}$. During inference, we sample for 100 steps with the guidance scale $\omega=7.5$.
More detailed settings can be found in the $Suppl$.

\noindent\textbf{Datasets.}
Following Custom Diffusion~\cite{CD}, we conduct experiments on twelve subjects selected from the DreamBooth dataset~\cite{DB} and CustomConcept101~\cite{CD}. They cover a wide range of categories including two scene categories, two pets, and eight objects.
\subsection{Comparative Study} \label{sec:Qualitative}
We conduct comparisons between our method and four existing methods: Textual Inversion (TI)~\cite{TI}, DreamBooth (DB)~\cite{DB}, Custom Diffusion (CD)~\cite{CD}, and Cones2~\cite{cones}.

\noindent\textbf{Qualitative comparison}. 
As shown in~\cref{fig:main comparison results}, TI and DB are limited to generating a single concept, whereas CD and Cones2 can produce multiple concepts but struggle with maintaining high fidelity. In contrast, our method excels in multi-concept generation, achieving both high image fidelity and prompt alignment, even in challenging long-format scenarios (third and fourth rows).

\begin{table*}[htbp]
\centering
  \tabcolsep=12pt
  \fontsize{9pt}{9pt}\selectfont
\resizebox{\linewidth}{!}
{
    \begin{tabular}{l|c c c|c c c|c|c}
    \toprule
      & \multicolumn{3}{c|}{\textbf{Single-Concept}} & \multicolumn{3}{c|}{\textbf{Multi-concept}} &  &  \\
      \multirow{-2}{*}{\textbf{Method}} & \rotatebox{0}{CLIP-I} & \rotatebox{0}{Seg CLIP-I} & \rotatebox{0}{CLIP-T} & \rotatebox{0}{CLIP-I} & \rotatebox{0}{Seg CLIP-I} & \rotatebox{0}{CLIP-T} & \multirow{-2}{*}{\shortstack[c]{\textbf{Training}\\ \textbf{Time} }} & \multirow{-2}{*}{\shortstack[c]{\textbf{Inference}\\ \textbf{Time} }} \\
     \midrule \midrule
TI\shortcite{TI} & 0.738 & 0.721 & 0.752 & 0.666 & 0.660 & 0.736 & 23min & \textbf{\color{blue}7.50s} \\ 
DB\shortcite{DB}        & \textbf{\color{blue}0.769} & 0.736 & 0.775 & 0.637 & 0.652 & \textbf{\color{blue}0.828} & 10min & \textbf{\color{red}7.35s} \\ 
Custom\shortcite{CD}  & 0.654 & 0.661 & \textbf{\color{red}0.813} & 0.624 & 0.637 & 0.812 & \textbf{\color{red}4min} & 7.53s \\ 
Cones2\shortcite{cones}            & 0.768 & \textbf{\color{blue}0.747} & 0.758 & \textbf{\color{blue}0.670} & \textbf{\color{blue}0.685} & 0.816 & 26min & 21.41s \\ \midrule
Ours              & \textbf{\color{red}0.783} & \textbf{\color{red}0.761} & \textbf{\color{blue}0.780} & \textbf{\color{red}0.714} & \textbf{\color{red}0.713} & \textbf{\color{red}0.838} & \textbf{\color{blue}6min} & 8.29s \\ 
 \bottomrule
 \end{tabular}
    }
    \caption{Quantitative comparisons. 
The best and second best results are in \textcolor{red}{red} and \textcolor{blue}{blue}, respectively. 
} 
  \label{tab:Quantitative}
\end{table*}

\noindent\textbf{Quantitative comparison}.
We assess all the methods using three evaluation metrics: CLIP-I, Seg CLIP-I, and CLIP-T.
(1) CLIP-I measures the average cosine similarity between the CLIP~\cite{clip} embeddings of the generated images and the source images. 
(2) Seg CLIP-I is similar to CLIP-I, but all the subjects in source images are segmented.
(3) CLIP-T calculates the average cosine similarity between the embeddings of prompt and image. 
As presented in~\cref{tab:Quantitative}, our method demonstrates superior image alignment and comparable text alignment in the single-concept setting. 
In the multi-concept setting, our method outperforms all the compared methods in the three selected metrics.
Moreover, with excellent image fidelity and prompt alignment ability, our method does not incur significant training and inference costs. 

\begin{figure}[h]
  \centering
  \includegraphics[width=\linewidth]{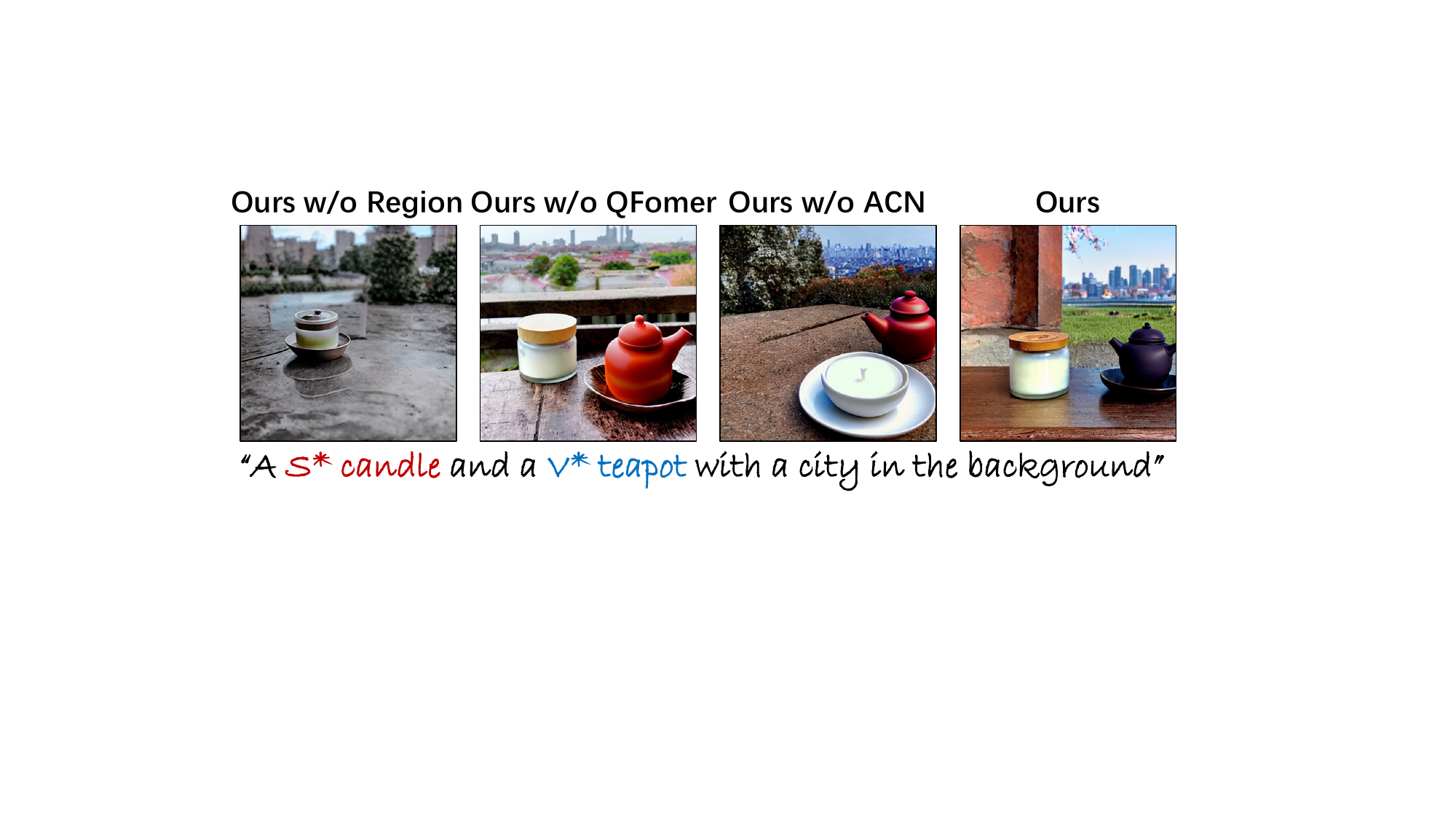}
  \caption{Qualitative ablation results. 
  }
  \label{fig:ablation}
\end{figure}

\subsection{Ablation Study} \label{sec:ablation}

\noindent\textbf{Regional Customization Module (RCM).} 
We first verify the effectiveness of RCM by simply removing it.
As shown in \cref{fig:ablation} and \cref{tab:ablation}, without RCM, the features of the candle and teapot have fused to some extent.
To further validate the effectiveness of  RCM, we retain our single concept learning (SCL) and replace our RCM with other layout T2I methods. 
We select two representative methods: LLM-grounded Diffusion (LG)~\cite{llm-ground} and BoxDiff~\cite{boxdiff}, with the bounding boxes used displayed on the left. 
On the one hand, LG~\cite{llm-ground} denoises each concept within the bounding boxes sequentially and then integrates them at the latent level, resulting in concept fusion in the overlapping regions. 
On the other hand, BoxDiff~\cite{boxdiff} employs the cross-attention map to construct a loss function for updating the latent variables. 
Although it can generate two concepts simultaneously, it suffers from low image fidelity.
Furthermore, neither of these methods can handle complex object interactions according to the given text prompt. 
In contrast, our method allows different single-concept modules to target specific regions at the cross-attention level, thereby generating multiple concepts simultaneously. 
By using a base prompt to guide complex object interactions across various regions, we can produce images with both high image fidelity and precise text alignment.
\begin{table}[htbp]
\centering
  \tabcolsep=5pt
  \fontsize{8pt}{8pt}\selectfont
{
    \begin{tabular}{l|c c c }
\toprule
     \textbf{Method} & \textbf{CLIP-I} & \textbf{Seg CLIP-I} & \textbf{CLIP-T} \\
    \midrule
    \midrule
    w/o Region & 0.691 & 0.707 & 0.710 \\
    w/o QFormer & 0.691 & 0.694 & 0.823 \\
    w/o ACN & 0.694 & 0.695 & 0.826 \\
    \midrule
    Ours & \textbf{0.713} & \textbf{0.712} & \textbf{0.838} \\
    
\bottomrule
    \end{tabular}
    }
    \caption{Quantitative ablation results.}
    \label{tab:ablation}
\end{table}

\noindent\textbf{QFormer and Adaptive Concept Normalization (ACN).}
We also demonstrate the effectiveness of the QFormer and the ACN by removing them either.
As shown in \cref{fig:ablation} and \cref{tab:ablation}, without QFormer or ACN, the fidelity of our method has decreased.
In contrast, our full method can faithfully perform multi-concept generation.

\subsection{Discussions} \label{sec:discuss}

\noindent\textbf{Inference time $\times$$N$ for $N$ concepts?} We also analyze the inference time of our method with the increasing number of concepts. As shown in~\cref{tab:inference}, the inference time of our method increases only slightly as the number of concepts grows.
This is because increasing concepts only leads to additional cross-attention computation in our RCM; other operations, like self-attention, residual addition, etc. remain the same as generating a single concept.

\begin{table}[h]
\centering
\tabcolsep=8pt
\fontsize{10pt}{9pt}\selectfont
\resizebox{\linewidth}{!}{
    \begin{tabular}{l|c c c }
\toprule
     & \textbf{2 Concepts} & \textbf{3 Concepts} & \textbf{4 Concepts} \\
    \midrule
    \midrule
    Inference Time & 8.29s & 10.07s & 10.53s \\
\bottomrule
    \end{tabular}}
\caption{Inference Time with more concepts.}
    \label{tab:inference}
\end{table}

\section{Conclusion}
We introduce MultiBooth, a novel and efficient framework for multi-concept customization (MCC). 
Compared with existing MCC methods, our MultiBooth allows plug-and-play multi-concept generation with high image fidelity while bringing minimal cost during training and inference.
By conducting qualitative and quantitative experiments, we demonstrate our superiority over state-of-the-art methods within diverse customization scenarios. We believe that our approach provides a novel insight for the community.

\section*{Acknowledgments}
This work was supported by the STI 2030-Major Projects under Grant 2021ZD0201404.

\bibliography{aaai25}

\end{document}